%% file: sample-xelatex.tex
\documentclass[sigconf,screen]{acmart}

\usepackage{multirow}

\def\BibTeX{{\rm B\kern-.05em{\sc i\kern-.025em b}\kern-.08emT\kern-.1667em\lower.7ex\hbox{E}\kern-.125emX}}

\acmSubmissionID{10}

 \setcopyright{acmcopyright}

\begin{document}
\fancyhead{}

\copyrightyear{2019}
\acmYear{2019}
\acmConference[MM '19]{Proceedings of the 27th ACM International Conference on
Multimedia}{October 21--25, 2019}{Nice, France}
\acmBooktitle{Proceedings of the 27th ACM International Conference on Multimedia (MM '19),
October 21--25, 2019, Nice, France}
\acmPrice{15.00}
\acmDOI{10.1145/3343031.3351093}
\acmISBN{978-1-4503-6889-6/19/10}

\title{You Only Recognize Once: Towards Fast Video Text Spotting}

\author{Zhanzhan Cheng$^{12*}$, Jing Lu$^{2*}$, Yi Niu$^2$, Shiliang Pu$^2$, Fei Wu$^{1+}$, Shuigeng Zhou$^3$}
\thanks{\textsuperscript{*}Both authors contributed equally to this research.}
\thanks{\textsuperscript{+}Corresponding author (wufei@cs.zju.edu.cn).}
\affiliation{%
  \textsuperscript{1}\institution{Zhejiang University, Hangzhou, China}
}
\email{chengzhanzhan@hikvision.com,wufei@cs.zju.edu.cn}
\affiliation{
  \textsuperscript{2}\institution{Hikvision Research Institute, China}
}
\email{lujing6,niuyi,pushiliang@hikvision.com}
\affiliation{
  \textsuperscript{3}\institution{Fudan University, Shanghai, China}
}
\email{sgzhou@fudan.edu.cn}

\input{sections/0_abstract.tex}

%

\ccsdesc[500]{Computing methodologies~Object detection}
\ccsdesc[500]{Computing methodologies~Object recognition}
\ccsdesc[500]{Computing methodologies~Tracking}
%
\keywords{video text spotting, detection, tracking, quality scoring}

%
\maketitle

\input{sections/1_intro.tex}
\input{sections/2_related.tex}

\input{sections/3_methods.tex}
\input{sections/4_dataset.tex}
\input{sections/5_exp.tex}

\input{sections/6_conclu.tex}

\bibliographystyle{ACM-Reference-Format}
\bibliography{egbib}

\end{document}

%% file: sections/0_abstract.tex
\begin{abstract}
  Video text spotting is still an important research topic due to its various real-applications.
  Previous approaches usually fall into the four-staged pipeline: text detection in individual images, frame-wisely recognizing localized text regions, tracking text streams and {generating} final results with complicated post-processing skills, { which} might suffer from the huge computational cost as well as the interferences of low-quality text.
  In this paper, we propose a fast and robust video text spotting framework by only recognizing the localized text one-time instead of frame-wisely recognition.
  Specifically, we first obtain text regions in videos with a well-designed spatial-temporal detector.
  Then we concentrate on developing a novel text recommender for selecting the highest-quality text from text streams and only recognizing the selected {ones}.
  Here, the recommender assembles text tracking, quality scoring and recognition into an end-to-end trainable module, which not only avoids the interferences from low-quality text but also dramatically speeds up the video text spotting process.
  In addition, we collect a larger scale video text dataset (LSVTD) for promoting the video text spotting community, which contains 100 text videos from 22 different real-life scenarios.
  Extensive experiments on two public benchmarks show that our method greatly speeds up the recognition process averagely by 71 times compared with the {frame-wise} manner, and also {achieves} the remarkable state-of-the-art.
\end{abstract}

%% file: sections/1_intro.tex
\section{Introduction}
Video text spotting is still an important research topic due to {the large} amount of video text reading applications such as
port container number identification in industrial monitoring and license plate recognition in intelligent transportation system.

\begin{figure}[t]
    \begin{center}
    \includegraphics[width=0.83\linewidth]{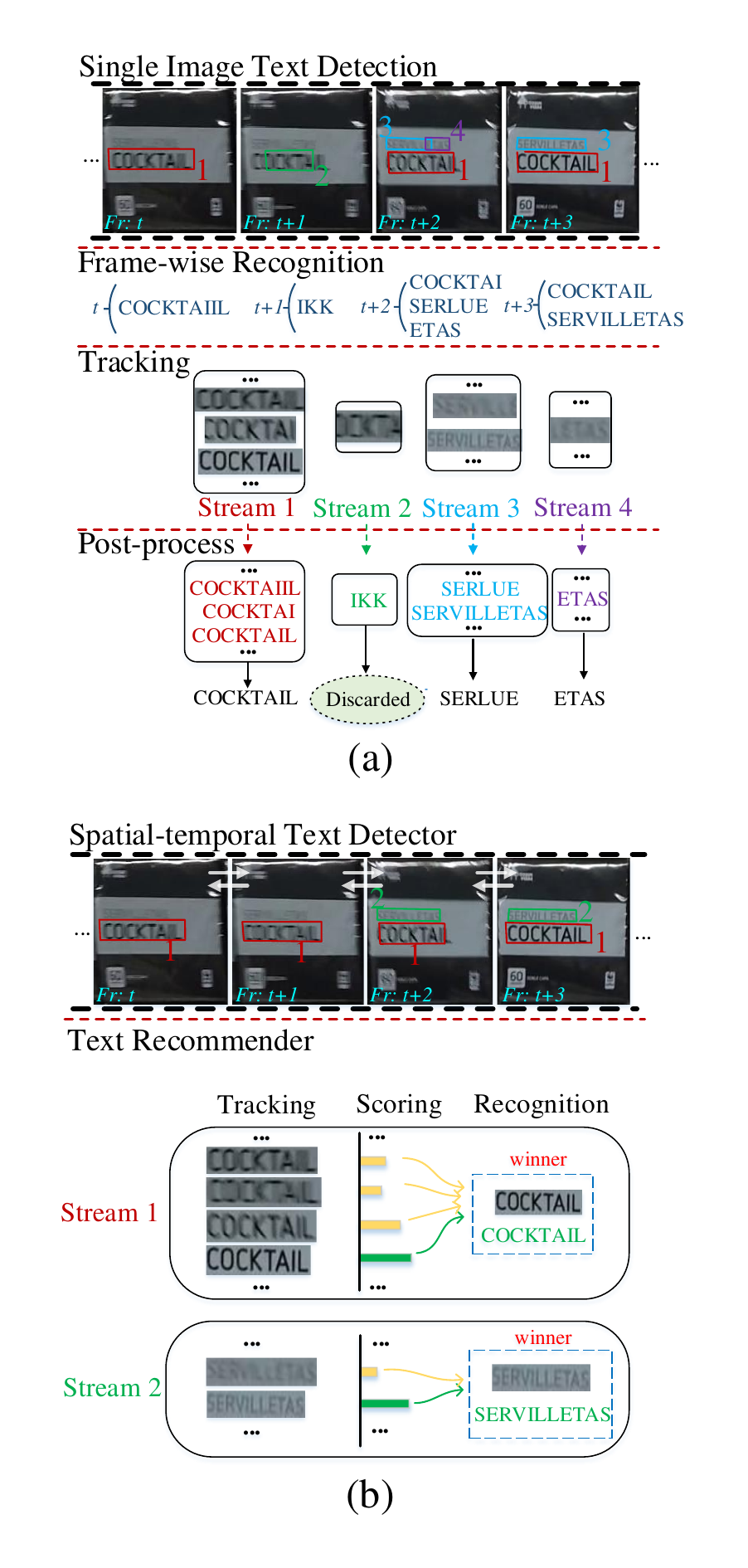}
        \vspace{-0.55cm}
    \end{center}
    \caption{
       Comparison of existing and proposed pipeline in video text spotting.
       (a) denotes the traditional framework which contains four steps: text detection in individual {frames}, frame-wisely recognizing text regions, tracking recognized text regions into streams and post-processing for obtaining final results.
       (b) denotes our proposed framework which only contains two steps: text detection among consecutive frames and recognizing the highest-quality text from text streams as the final results.
        $Fr\text{:} (.)$ means the frame ID, and the numbers {next to} their corresponding boxes in each frame mean the tracked IDs.
        The white arrows in (b) {represent} the temporal relation built between consecutive frames in detection process.
        [Best viewed in color]
    }
    \vspace{-0.45cm}
    \label{fig-framework}
\end{figure}

In the past few years, several  methods \cite{merino2014real,nguyen2014video,wang2017end} are proposed to read video scene text with a multi-stage pipeline strategy: that is,
firstly localizing text regions in individual frames, recognizing localized text regions one-by-one, tracking recognized regions as text streams, and applying post-precessing techniques for generating the final results. Figure~\ref{fig-framework}.(a) shows such process.

However, above methods still suffer from two problems:
1) Excessive computational cost for directly reading text regions one-by-one, which may be impractical especially when working on front-end devices such as surveillance video cameras.
2) Much unstable recognition results due to the overwhelming of \emph{low-quality} (\emph{eg.} blurring, perspective distortion, rotation and poor illumination etc.) text.
For example, the low-quality text region in 'stream 3' of Figure \ref{fig-framework}.(a) leads to the wrong result 'SERLUE'.
In practical, it is unnecessary to recognize each text region in text streams, which will result in huge {time cost} and also {bring} various interferences of low-quality text.

To circumvent these problems, an idea is to select the \emph{highest-quality} (\emph{eg.} clear and horizontal) text {region} from each text stream to be recognized once instead of one-by-one, which needs us design a robust text quality scorer to assign a quality score for each candidate text region, as shown in Figure \ref{fig-framework}.(b).
For further speeding up the process of video text spotting, we also attempt to assemble the text tracking, text quality scoring and recognition into an end-to-end trainable module (denoted by \textbf{text recommender}).
The text recommender will benefit from the complementarity among tracking, scoring and recognition, and then {improve} the text spotting performance again.

In this paper, we propose a fast and robust video text spotting approach named as YORO by combining a well-designed spatial-temporal video text detector and a novel text recommender into a whole framework.
Concretely,
the spatial-temporal detector is designed to recall more text by referring to temporal relationship among different frames.
The text recommender assembles the text tracking for generating text streams, the text quality scoring {for assessing the quality of} each text region and the recognition into an end-to-end trainable network, as shown in Figure \ref{fig_our_frame} (b), in which the quality scores are learnt {in a weakly-supervised manner} (detailed in Methodology Section).
In this way, we can ignore low-quality text regions and only reserve highest-quality text regions for recognition, which not only improves the recognition performance but also greatly decreases the computational cost.

Last but not least, we also note that the scales of existing video text benchmarks are very limited.
For example, the largest video scene text dataset `Text in Videos' \cite{karatzas2015icdar} 
only contains 49 videos 
from 7 different scenarios, which may limit the video text recognition research progress.
Therefore, we collect a \textbf{l}arger-\textbf{s}cale \textbf{v}ideo \textbf{t}ext \textbf{d}ataset (\emph{abbr}. LSVTD) containing 100 videos from 22 natural scenarios, and hope to help the research of video text understanding.

Main contributions of this paper are as follows:
(1) We design a novel text recommender for selecting the highest-quality text from text streams and then only recognizing the selected text {regions} once, which significantly speeds up the recognition process, and also {improves} the video text spotting performance.
(2) We integrate a well-designed spatial-temporal text detector and {a} text recommender into {an} unified two-stage framework YORO for fast end-to-end video text spotting.
(3) In order to promote the progress of video text spotting, we collect and annotate a larger scale video text dataset, which contains 100 videos from 22 different real-life situations.
(4) Extensive experiments demonstrate that our method is fast and robust and achieves impressive performance in video scene text reading.

%% file: sections/2_related.tex
\section{Related work}
\subsection{Text Reading in Single Images}
Traditionally, the scene text reading system contains a {text detector} for localizing each text region and a {text recognizer} for generating corresponding character sequences.
For text detection, numerous methods are proposed to localize regular and irregular (oriented and curved etc.) text regions, which can be categorized as anchor-based \cite{hu2017wordsup,jiang2017r2cnn,liao2018rotation,liu2017deep,ma2018arbitrary,shi2017detecting} and direct-regression-based \cite{he2017deep,wang2018geometry,zhou2017east}.
For text recognition, the task is now treated as a sequence recognition problem, in which CTC \cite{Graves2006}-based \cite{borisyuk2018rosetta,shi2017end,wang2017gated} and attention-based \cite{bai2018ep,cheng2017focusing,cheng2018aon,shi2016robust,shi2018aster} methods are designed and have achieved promising results.
Recently, there are several methods \cite{borisyuk2018rosetta,he2018end,li2017towards,liu2018fots} attempting to spot text end-to-end. 

In fact, lots of text reading applications actually work in video scenarios, in which scene text spotting from multiple frames may be more meaningful.

\subsection{Text Reading in Videos}
In recent years, only a few attention has been drawn to spotting video scene text in contrast to {text reading} in still images.
For more details of text detection, tracking and recognition in video, the readers can refer to a comprehensive survey \cite{yin2016text}.
In general, reading text from scene videos can be roughly categorized into three major modules: 1) text detection, 2) text tracking, and 3) text recognition. 

\textbf{Text detection in videos.}
In early years (before 2012), most of methods focus on detecting text in each frame with connected component analysis \cite{yin2013robust} 
or
sliding window \cite{kim2003texture} strategy. 
However, the performance of them is limited due to the low representation of handcrafted features.
Though the recent detection techniques (mentioned in \emph{Section}. 2.1) in still images can help improve feature representation, detecting text in scene videos is still challenging because of its {complicated} temporal characteristics (eg. motion).
Therefore, text tracking strategies are introduced for enhancing the detection performance, which are further divided into two categories \cite{yin2016text}: spatial-temporal information based methods
\cite{goto2009text,shiratori2006efficient,tanaka2007autonomous,tanaka2008text} 
for reducing noise and fusion based methods
\cite{gomez2014mser,minetto2011snoopertrack,zuo2015multi} 
for improving detection accuracy.
Recently, Wang et al. \cite{wang2018scene} employed optical flow based method to refine text locations in the subsequent frames.

\textbf{Text tracking in videos.}
The traditional methods such as template matching
\cite{fragoso2011translatar,na2010effective,shiratori2006efficient,tanaka2007autonomous} 
and particle filtering were popular.
But these methods failed to solve the {re-initialization} problem, especially in scene videos.
Then the tracking-by-detection based methods
\cite{nguyen2014video,petter2011automatic,rong2014scene} 
were developed to estimate the tracking trajectories and solve this problem.

Recently, Zuo et al. \cite{zuo2015multi} and Tian et al. \cite{tian2016scene} attempted to fuse multi-tracking strategies (eg. spatial-temporal context learning \cite{zhang2014fast}, tracking-by-detection etc.) for text tracking, in which Hungarian \cite{kuhn1955hungarian} algorithm was applied for generating the final text streams.
Yang et al. \cite{yang2017unified} also proposed a motion-based tracking approach in which detected results are directly propagated to the neighboring frames for recovering missing text regions.
In fact, the robust feature extractor is the most important component for a text tracker. 

\textbf{Text recognition in videos.}
With the tracked text streams, there are two strategies for better scene text recognition: selection strategy by selecting the best text regions from streams (popular before 2010), and results fusion strategy by combining corresponding recognized character results.
Correspondingly, methods
\cite{shiratori2006efficient,tanaka2007autonomous,tanaka2008text} 
selected the region with the longest horizontal length as the most appropriate region. 
Then Goto and Tanaka
\cite{goto2009text} 
further enhanced the selection algorithm by taking six different features (eg. Fisher's discriminant ratio, text region area etc.) into account.
While recent methods \cite{greenhalgh2015recognizing,rong2014scene} directly fused recognized results in text streams for final text prediction by majority voting, CRF or frame-wise comparison, and these approaches assumed that recognition results in most frames are trust-worthy, which may not be true in unconstrained scenarios.
In addition, frame-wise text recognition also results in high computation cost.

\textbf{End-to-end text recognition in videos.}
There are several works proposed to solve the end-to-end video text spotting problem.
Nguyen et al. \cite{nguyen2014video} first proposed an end-to-end video text reading solution by extending Wangs's method \cite{wang2011end}, in which the frame-wise detection and the tracking with multiple features (eg. the temporal distance, edit distance etc.) are applied. 
Merino-Gracia and Mirmehdi \cite{merino2014real} proposed an end-to-end video scene text reading system by introducing the unscented Kalman filter
\cite{wan2000unscented}
, but {mainly} focused on large text found in outdoor environments.
Recently, {Wang et al. \cite{wang2017end} firstly utilized an end-to-end deep neural network to detect and recognize text in each frame, and then employed the \emph{tracking-by-detection} strategy to associate text regions, and recovered missed detections with the tracked results, finally obtain the recognition results by voting the most frequently appeared text strings.}

Different from frame-wise recognition in \cite{merino2014real,nguyen2014video,wang2017end},
in this paper we propose {a} fast and robust video scene text spotting framework by integrating a spatial-temporal video text detector and a novel text recommender, which {dramatically} speeds up the text spotting process and also improves the final recognition performance.

%% file: sections/3_methods.tex
\section{Methodology}
\begin{figure*}
\begin{center}
\vspace{-0.5cm}
\includegraphics[width=1.\linewidth]{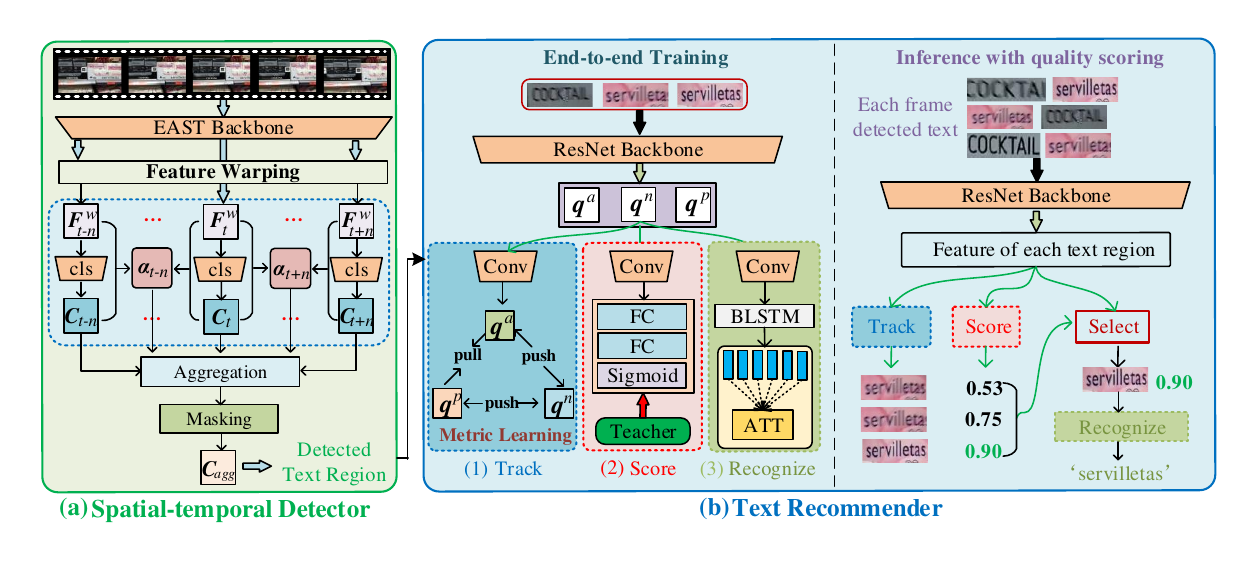}
\vspace{-0.85cm}
\end{center}
   \caption{
    The workflow of YORO, which consists of two submodules:
    (a) The spatial-temporal text detector for generating text regions;
    (b) The text recommender assembling the tracking, quality scoring and recognition into a whole end-to-end trainable module, in which the left and right parts separately denote the training and inference processes. Specifically, the tracking, scoring and recognition are simultaneously learnt with the metric learning, the teacher-student strategy and the attention mechanism, respectively.
   }
   \vspace{-0.3cm}
\label{fig_our_frame}
\end{figure*}
\subsection{Overview}
The architecture of YORO is shown in Figure \ref{fig_our_frame}, which consists of two modules:
the spatial-temporal video text detector for detecting text regions among adjacent frames
and the text recommender for recognizing the highest-quality text regions from the tracked text streams.

Concretely,
the video text detector is responsible for recalling more text by referring to temporal relationship among different frames.
The recommender is developed with three submodules:
the essential quality scoring component for directly generating the quality score of each text region by borrowing the `teacher-student' concept \cite{hinton2015distilling}, the tracking component for generating text streams with metric learning techniques which can enhance and simplify the tracking process,
and the recognition component for generating the final results with {an} attention-based decoder \cite{shi2016robust}.
Note that, the recommender is implemented as {an} end-to-end trainable network.

\subsection{Spatial-Temporal Detection}
The text detection architecture is shown in Figure \ref{fig_our_frame}.(a), in which the backbone of EAST \cite{zhou2017east} is selected as our backbone.
Here, we learn relations between consecutive frames with a \emph{spatial-temporal aggregation} strategy for improving video text detection process, which can be divided into three steps: 1) enhancing temporal coherence between frames with a feature warping mechanism \cite{zhu2017flow}, 2) spatial matching between frames with a comparing and matching strategy inspired by \cite{chen2018video,si2018dual}, and 3) temporal aggregation.

Formally,
let $I_t$ be the $t\text{-}th$ frame in a video, the detection results in $I_t$ can be refined with the {detections} of its consecutive frames $(I_{t-n}, ..., I_{t+n})$ where the size of refining window is $2n\text{+}1$.

\emph{Enhancing temporal coherence}.
We obtain the corresponding sequence of feature maps $F\text{=}(F_{t-n}, ..., F_{t+n})$ by propagating frames through the EAST backbone.
Given a pair of frame features $F_{t+i}$ and $F_t$ (the reference frame), we enhance their temporal coherence by referring to the estimated flow $flow_{(t+i, t)}$ between $I_{t+i}$ and $I_t$ with a flow-guided warping mechanism
\begin{equation}
F_{t+i}^w = Warp(F_{t+i}, flow_{(t+i, t)}),
\end{equation}
where $flow_{(t+i, t)}$ is pre-computed with TV-L1 algorithm, $Warp(.)$ is the bilinear warping function applied on each {element} in the feature maps, and $F_{t+i}^w$ denotes the feature maps warped from frame $I_{t+i}$ to frame $I_t$.
Thus $F$ is further transferred as the warped $F^w = (F_{t-n}^w, ..., F_{t+n}^w)$.
Then we generate an enhanced sequence of \emph{confidence maps} $C=(C_{t-n}, ..., C_{t+n})$ by propagating $F^w$ into a classification sub-network, in which each value in $C_{t+i}$ represents the possibility of being a text region.

\emph{Comparing and matching}.
We evaluate the spatial matching degree of two frames with matching weights.
The weights are firstly computed with a transform module to produce the feature-aware filter  
which is represented as
\begin{equation}
F_{t+i}^{trans} = ReLU(BN(W F_{t+i}^w + b)),
\end{equation}
where $W$ and $b$ are learnable parameters, BN and ReLU represent Batch Normalization and rectified linear unit function, respectively.
Given the transformed feature maps, 
we compute the similarity energy $Sim_{t+i,t}=F_{t+i}^{trans} \odot F_{t}^{trans}$ of $I_{t+i}$ and $I_t$ as the matching weights, where $\odot$ means the dot product position-wisely.

\emph{Temporal aggregation}.
Then we compute the aggregation weights by
\begin{equation}
a_{t+i} = \frac{exp(Sim_{t+i,t} \odot C_{t+i})}{ \sum_{i^\prime=-n}^{n} exp(Sim_{t+i^\prime,t} \odot C_{t+i^\prime}) }.
\end{equation}
Here, we multiply $Sim_{t+i,t}$ by $C_{t+i}$ in order to reinforce the aggregation weights of positive detections. 

Finally, the temporal aggregation across the consecutive frames is computed by
\begin{equation}
C_{t, agg} = \sum_{i=-n}^{n}a_{t+i} * C_{t+i}
\end{equation}
where "*" represents element-wise production.

For handling few mis-aggregated situations, we further refine $C_{t,agg}$ as $C_{t, ref} = C_{t, agg}*M_{t}$ by applying a normalized binary mask $M_t$ to $C_{t,agg}$,
where $M_t$ is calculated by normalizing $F_t^w$ as a binary mask with a pre-set threshold (default by 0.5).
The detailed training process is {the} same to that in EAST \cite{zhou2017east}.

\subsection{Text Recommender}
The text recommender contains three trainable parts: the quality scoring, tracking and recognition, as shown in Figure \ref{fig_our_frame}.(b).
\subsubsection{Text Quality Scoring}
It is almost impossible to manually annotate the quality score for each text region because the judgement of imaging quality is subjective for different annotators, and the annotating cost is also very tremendous.
Therefore, we design {an} off-line scoring strategy (denoted by teacher) for generating the quality score labels {in a weakly-supervised fashion}, and then train the lightweight quality scoring network (denoted by student) with the teacher's scores.
\begin{figure}[h]
\begin{center}
   \includegraphics[width=1\linewidth]{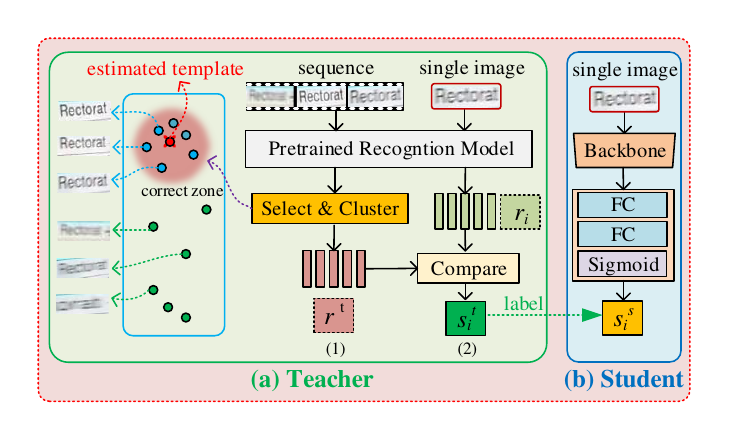}
   \vspace{-0.55cm}
\end{center}
   \caption{Training mechanism of quality scoring module.
   (a) Teacher: the off-line scoring mechanism to generate quality score labels by referring the estimate {templates}.
   Here, different quality of text regions are {located at different positions in feature space}, in which those always recognized correctly will fall into a `correct zone' and its center {corresponds} to the estimated position of standard template.
   (b) Student: the quality scoring network.
   }
\vspace{-0.45cm}
\label{fig_qsn}
\end{figure}

\emph{Teacher}.
For teaching the student module, the teacher module should generate explicit quality score for each text region.
Here, we propose the concept of \emph{standard template}, which means that the template is the feature representation of an ideal text image without any interference.
Then each text image is corresponding to a standard template image in the feature space, and we can define the distance between the template and a candidate text region as the corresponding quality score.
In this assumption, those images close to its template are more possible to be correctly recognized, vice versa.

However, it's unrealistic to directly calculate the distance between text regions in the low-level feature space (\emph{eg.} color and intensity) because it is hard to directly align two text regions due to the {difference} on color, intensity and scales etc. 
Then, we attempt to represent each text region with the extracted {high-level} character features. For example, an image with content `ABC' will be represented as the concatenation of high-level feature `A', `B' and `C'.
Here, the character features can be provided by the recognition model such as the output features of the attention-based decoder.

Specifically, we first feed a text stream 
into a pre-trained recognition model, and then represent each of them with the concatenated character features from the recognition model.
We empirically find that features of correctly recognized text tend to be localized in a compact area while the features of wrongly recognized ones are usually far from this compact region, as shown in Figure \ref{fig_qsn}.
Therefore, we conduct the estimation of the template image $r^t$ with the correctly recognized feature sets $\{r^{cor}_1,r^{cor}_1,...,r^{cor}_k\}$ by applying the K-Means algorithm:
\begin{equation}
{r}^{t} \text{=} kmeans(r^{cor}_1,r^{cor}_1,...,r^{cor}_k),
\label{L_score}
\end{equation}
Consequentially, we calculate the feature distance between each text region and its estimated template image as the quality score:
\begin{equation}
{s}^{t}_{i} \text{=} \frac{r^t \odot r_i}{||r^t|| * ||r_i||},
\label{L_score}
\end{equation}
where $\odot$ means the dot product and $||.||$ means the Euclidean distance.

\emph{Student}.
Supervised by the generated quality scores from teacher, the quality scoring network can directly regress the quality score for each text region, as illustrated in Figure \ref{fig_qsn}.(b).
Then the loss function for quality scoring network is as follows:
\begin{equation}
\mathcal{L}_{S} \text{=} \frac{1}{N}\sum_{i=0}^{N}||s^t_i\text{-}s^s_i||
\label{L_score}
\end{equation}
where $s^t$ is the quality score generated by teacher, and $s^s$ is the regressed quality score by student. 

\subsubsection{Text tracking}
The tracking task aims to group corresponding text regions into text streams, shown in the left part of Figure \ref{fig_our_frame}.(b).
Intuitively, the tracker should have the ability to ensure that the features of a text region in one stream {must remain} closer distance to those in the same stream than others, which implies:
1) the features must be discriminative enough to tolerate various interferences in unconstrained scenes, and 2) the module may be better if trained with a good distance measure.

\emph{Robust feature extraction}.
Thanks to the studies in deep neural network and metric learning, we extract robust features for the tracker by applying metric learning technique.
Concretely,
we firstly select three regions from localized candidate regions as an image triplet $(R^a, R^p, R^n)$, in which $R^a \text{and} R^p$ are corresponding to the same text instance while $R^n$ is randomly selected from other text instances.
Secondly, an image triplet is fed into a deep CNN for generating its L2 Normalized high-level representation $(q^a, q^p, q^n)$.
The tracker is {then} trained with two metric learning loss: contrastive loss \cite{hadsell2006dimensionality} $\mathcal{L}_{contra}$ and triplet loss \cite{schroff2015facenet} $\mathcal{L}_{triplet}$, i.e.
\begin{equation}
\mathcal{L}_{T} = \mathcal{L}_{contra} + \lambda_t \mathcal{L}_{triplet},
\label{loss_tssn}
\end{equation}
where $\lambda_t$ is default by 1.

\emph{Text stream generation}.
With the trained tracking model,
for a pair of candidate text regions $(R^1, R^2)$, we calculate its matching cost by
\begin{equation}
MC(R^{1},R^{2}) = \frac{1}{q^1 \odot q^2+\epsilon}.
\label{match-cost}
\end{equation}
For avoiding division by zero error, $\epsilon$ is set as $10^{-7}$. 
Then those pairs with $MC$ larger than a threshold are considered as invalid matching pairs and filtered out.
Finally, we employ Hungarian algorithm \cite{kuhn1955hungarian} to generate the text streams.

\subsubsection{Text Recognition}
The text recognition module is not our focus, and we select attention-based method as our decoder just like in \cite{cheng2017focusing,shi2016robust,shi2018aster}.
\subsubsection{End-to-end training}
Since the above three submodules are complementary for extracting discriminative features, we jointly train all the three tasks with same backbone in an end-to-end trainable network, which is supervised by:
\begin{equation}
\mathcal{L} = \lambda_{1}\mathcal{L}_{T}+\lambda_{2}\mathcal{L}_{S}+\lambda_{3}\mathcal{L}_{R},
\label{joint}
\end{equation}
where $\mathcal{L}_{R}$ is the recognition part loss and $\lambda$$_{i}$ (\emph{i}=1,2,3) denotes the loss weight for different tasks.
Settings are detailed in \emph{Experiment} section.

%% file: sections/4_dataset.tex
\section{The Larger-Scale Video Text Dataset}
In recent years, research in video scene text still remains unpopular in contrast to its promising application prospect.
The existing video scene text datasets such as \emph{IC13} or \emph{IC15} (detailed in Table \ref{tab_dataset_scenes}) are limited on the scale of video items and scenarios, which may restrain research of video scene text spotting.
Therefore, we collect and annotate a \textbf{l}arger-\textbf{s}cale \textbf{v}ideo \textbf{t}ext \textbf{d}ataset (LSVTD), which contains 100 scene videos acquired from 22 typical real-life scenarios.
\begin{table}[h]
    \begin{center}
    \scalebox{0.8}{
    \begin{tabular}{|l|c|c|c|c|c|}
    \hline
     Datasets&                              \#scenarios & \#videos & \#frames & \#instances & quality? \\
    \hline
    Merino \cite{merino2007framework}       & 4 & -- & -- & -- &   \\
    Minetto \cite{minetto2011snoopertrack}  & -- & 5 & 3599 & 8706 &   \\
    IC13 \cite{karatzas2013icdar}           & 7 & 28 & 15277 & 93934 & \checkmark\\
    YVT \cite{nguyen2014video}           & -- & 30 & 13500 & -- & \\
    IC15 \cite{karatzas2015icdar}           & 7 & 49 & 27824 & -- & \checkmark \\ \hline
    LSVTD                                   & \textbf{22} & \textbf{100} & \textbf{66700} & \textbf{569300} & \checkmark \\
    \hline
    \end{tabular}
    }
    \end{center}
    \caption{
        Comparison with existing video text datasets.
        `\emph{quality?}' means if each text region is labeled with quality score.
    }
    \vspace{-0.45cm}
\label{tab_dataset_scenes}
\end{table}

\textbf{Dataset characteristics}.
LSVTD is detailed in Table ~\ref{tab_dataset_scenes}, and mainly characterized by
1) Much larger scale, which is more than twice the scale of IC15.
2) More diversified scenarios. LSVTD covers a wide range of 13 indoor (eg. bookstore, shopping mall) and 9 outdoor (eg. highway, city road) scenarios, which is more than 3 times the diversity of IC15.
The variety of scenarios challenges text spotting algorithms to achieve robust performance.
3) Multilingual text instances.
LSVTD contains text with more than three kinds of languages (English, Spanish and Chinese etc.), which are further divided into 2 major categories: Latin and Non-Latin.
We additionally label this attribute for the convenience of evaluation {for} different algorithms.
More details of LSVTD {are} shown in the {supplementary materials}\textsuperscript{1}.

\textbf{Annotation details}.
Following the annotation details adopted by IC15, we annotate the following attributes for each text:
1) \emph{Polygon coordinate} represents text location;
2) \emph{ID} means the unique identification for each text among consecutive frames; That is, the same text in consecutive frames shares the same ID;
3) \emph{Language} is categorized as Latin and Non-Latin as mentioned above;
4) \emph{Quality} indicates the image quality of each text region, which is labelled as `high',`moderate' or `low'.
5) \emph{Transcripts} means text string for each text region.

These videos are collected by our 6 colleagues and {range} from 5 seconds to 1 minute.
It took more than 3500 man-hours for annotating these frames.
Here, we hope that the availability\footnote{Available at https://davar-lab.github.io/opensource/dataset/lsvtd} of LSVTD may spur more interests in corresponding areas. 

%% file: sections/5_exp.tex
\section{Experiments}
\subsection{Evaluation Protocols}
The evaluation protocols for text detection, tracking and recognition in videos have been declared in \cite{yin2016text}. We only select several mainstream evaluation metrics in this paper.

\textbf{Detection metrics}.
Following detection methods \cite{jiang2017r2cnn,zhong2018anchor,zhou2017east}, \emph{precision} (denoted by PRE), \emph{recall} (denoted by REC) and \emph{F-measure} are selected as the evaluation metrics.

\textbf{Tracking metrics}.
The tracking metrics should maximize the sum of overlap between the tracking results and ground truth.
In general, multiple object tracking precision (\emph{abbr}. MOTP), multiple object tracking accuracy (\emph{abbr}. MOTA), and the average tracking accuracy (\emph{abbr}. ATA) are used to evaluate the performance of tracking.

\textbf{Quality scoring metrics}.
Naturally, the better quality frames are selected, the higher recognition accuracy we will get.
To evaluate the performance of the quality scoring mechanism, we first define the \textbf{q}uality \textbf{s}election \textbf{h}itting \textbf{r}ate(\emph{abbr}. QSHR) for evaluating the selection accuracy
\begin{equation}
QSHR \text{=} \sum_{i=0}^{N}\frac{\bar{q}_{i}}{N},
\end{equation}
where $N$ denotes the number of text streams, and $\bar{q}_i\in\{0, 1\}$.
In the $i\text{-}th$ text stream, $\bar{q}_i\text{=}1$ means the region annotated with "high" is hit, 0 otherwise.

Based on the selection mechanism, we further define the \textbf{r}ate of \textbf{c}orrectly \textbf{r}ecognizing selected text regions (\emph{abbr}. RCR) for evaluating sequence-level recognition accuracy
\begin{equation}
RCR \text{=} \sum_{i=0}^{N}\frac{\bar{a}_{i}}{N},
\end{equation}
where $\bar{a}_i\in\{0, 1\}$.
In the $i\text{-}th$ text stream, $\bar{a}_i\text{=}1$ means the selected text region is correctly recognized, 0 otherwise.

\textbf{End-to-end metrics}.
In previous methods, MOTP, MOTA and ATA are generally used in end-to-end evaluation, which {evaluate} performance in {word-level recognition}. That is, a predict word is considered as a true positive if its IoU over ground truth is larger than 0.5 and the word recognition is correct.

However, in our task, we just score and recognize a tracked text stream one-time.
According to the selection-and-recognition strategy, we redefine the end-to-end metrics by considering two constrains:
1) The recognized result of selected region should match the corresponding text transcription. 
2) The temporal location (frame ID) of selected region should fall into the interval between the annotated starting and ending frame. 
In addition, the selected candidate should have a spatial overlap ratio (default by over 0.5) with the annotated bounding box.
Thus we define the {sequence-level recall $REC_s$ and precision ($PRE_s$)}
\begin{equation}
REC_s = \frac{N_{r}}{N_{g}}, ~~~PRE_s=\frac{N_{r}}{N_{d}},
\end{equation}
for \emph{constrain 1} and \emph{constrain 2}, in which
${N}$$_{r}$, ${N}$$_{g}$ and ${N}$$_{d}$ separately denote the number of valid recalled streams, the number of total ground truth streams and the number of detected text streams.
Correspondingly, the {sequence-level} F-score ($F\text{-}Score$) is denoted as
\begin{equation}
F\text{-}Score=\frac{2\ast{PRE_s}\ast{REC_s}}{PRE_s+REC_s}
\end{equation}
by simultaneously considering $PRE_s$ and $REC_s$.

It's worth to note that we only match a given ground truth stream once, which also {penalize} the stream fragmentation problem occurred in text tracking.
In all, the evaluation protocol measures the accuracy and efficiency to extract useful text information from videos.

\subsection{Implementation Details}
All of our work is built on the CAFFE framework.

\emph{Detection Network}.
The EAST backbone is pre-trained on the `Incidental Scene Text' dataset \cite{karatzas2015icdar} and `COCO-Text' dataset \cite{veit2016cocotext} by following \cite{liu2018fots},
and then the model is fine-tuned on corresponding video training set such as  IC13 or IC15.
Images are randomly cropped and resized to 512$\times$512 and then fed into the network.
In training stage, we set \emph{batch-size}=4 and learn the network by adopting `Adam' with \emph{learning rate=$10^{-4}$, decay rate=0.94} for every $10^4$ iterations, in which text regions with short side less than 10 pixels are ignored during training.
While in testing stage, we only conduct the single-scale testing.
In the post-processing stage, we adopt NMS on predicted geometric shapes with \emph{threshold=0.2}.

\emph{Text Recommender Network}.
The `ResNet Backbone'+`Conv Blocks' used in tracking, scoring and recognition is adopted from an image encoder used in \cite{cheng2017focusing}, and the `BLSTM'+`ATT' module in scoring and recognition is an attention decoder used in \cite{cheng2017focusing,shi2016robust}.
The network is pre-trained on the 8-million synthetic data \cite{jaderberg2014synthetic} using `Adadelta' by following \cite{shi2016robust}, and further fine-tuned on IC13 or IC15 using SGD with the fixed learning rate of $10^{-4}$.
The loss weights $\lambda$$_{1}$, $\lambda$$_{2}$, $\lambda$$_{3}$ in Equation. \ref{joint} are all set to 1, and the margin $\alpha$ used in triplet loss is set to 0.8.
In text tracking process, the threshold $MC$ for filtering out invalid text pairs is set to 0.92.

\subsection{Ablation Study}

\textbf{Effects of text recommender.}
In IC13 and IC15, text regions are annotated as 3 quality levels (`low', `moderate' and `high').
Those streams containing at least two types of quality annotations are treated as our testing dataset.

To evaluate the proposed text recommender (\emph{abbr.} TR), we compare our method with two commonly used {selection-and-recognition} strategies:
1) Using the predicted confidence (the average probability of generating characters) of a word as the quality score (denoted by PCW).
2) Selecting the text region with the highest frequency of predicted results as the voted best one (denoted by HFP), which is similar to the \emph{majority voting} strategy used in \cite{wang2017end}.
\begin{table}[h]
\begin{center}
\scalebox{1}{
\begin{tabular}{|l|c|c|c|}
\hline
Methods & $QSHR$ & $RCR$  & \multirow{2}{*}{$FPS$}  \\
        & (IC13/IC15) &  (IC13/IC15) & \\
\hline
PCW & 74.55/75.83 & 66.06/66.32 & \multirow{2}{*}{4.52}\\             
HFP & 75.32/76.34 & 68.30/68.56 & \\ \hline
TR ($\mathcal{L}_{S}$) & 77.89/79.69 & 68.89/69.41  & \multirow{4}{*}{324.58}\\         
TR ($\mathcal{L}_{S}$+$\mathcal{L}_{T}$) & 78.64/80.36 & 69.12/69.82  & \\
TR ($\mathcal{L}_{S}$+$\mathcal{L}_{R}$) & 81.23/83.03 & 69.92/70.69  & \\                                              
TR ($\mathcal{L}$) & \textbf{81.74}/\textbf{83.29} & \textbf{70.18}/\textbf{70.95}  & \\             
\hline
\end{tabular}
}
\end{center}
\caption{Effects of TR on IC13 and IC15 compared with other frame selection methods. (.) refers to the loss applied in TR. 
}
\label{tab_fsn}
\end{table}

Table. \ref{tab_fsn} shows the results. 
Compared to HFP, TR ($\mathcal{L}$) significantly {improves} the \emph{QSHR} performance by 6.4\% on IC13 and 7\% on IC15, while improves the \emph{RCR} performance by 1.9\% on IC13 and 2.4\% on IC15.
Compared to the results of other settings (TR ($\mathcal{L}_{S}$), TR ($\mathcal{L}_{S}$+$\mathcal{L}_{R}$) and TR ($\mathcal{L}_{S}$+$\mathcal{L}_{T}$)), the end-to-end training strategy can further boosting the scoring performance.
For {simplified} representation, we use TR as TR ($\mathcal{L}$) in the sequel.

Moreover, TR only needs recognize a text stream one-time, which can greatly decrease the computational cost. As shown in Table. \ref{tab_fsn}, TR speeds up the recognition process averagely by 71 times compared with the frame-wise manner on two datasets.

\textbf{Extreme testing for text recommender.}
It is worth noticing that TR can still select the best one when handling text streams with a large proportion of low-quality text regions, while the voting strategy becomes useless. It implies that TR is more robust in complex and heavily distorted video scenarios.
Therefore, we conduct extreme testing on a constituted \emph{low-quality text stream set} by discarding all streams containing more than 40\% highest quality text regions on IC13 and IC15.
We calculate the \emph{QSHR} and \emph{RCR} on this set by checking whether the highest quality of text is hit and whether the selected text is correctly recognized.
Table \ref{tab_tssn} gives the results and demonstrates that TR is more robust in complex and low-quality video scenarios.
\begin{table}[h]
\begin{center}
\scalebox{1}{
\begin{tabular}{|l|c|c|}
\hline
 Methods & $QSHR$ & $RCR$  \\
 & (IC13/IC15) &  (IC13/IC15)  \\
\hline
 PCW & 41.73/45.66 & 59.78/60.62  \\
 HFP & 39.37/41.73 & 58.96/60.06  \\
 TR & \textbf{51.18}/\textbf{54.33} & \textbf{66.14}/\textbf{67.71}  \\         
\hline
\end{tabular}
}
\end{center}
\caption{Extreme testing of TR on IC13 and IC15 compared with other frame selection methods.
}
\label{tab_tssn}
\end{table}

\textbf{End-to-end evaluation.}
To analyze the contributions of above components, we conduct the end-to-end evaluation on the popular IC15 dataset, as shown in Table. \ref{end-our}.

For fair comparison, we here declare some baselines in our framework, i.e.
For detection, we treat the video text detection without our spatial-temporal strategy as our detection baseline (denoted by D-BASE), and denote the spatial-temporal detection model as D-ST;
While for text recommender, compared with the trainable TR, we treat TR ($\mathcal{L}_{S}$) as our baseline.

From Table. \ref{end-our}, we find that
\begin{itemize}
  \item Comparing to the detection baseline, the spatial-temporal strategy can steadily help the end-to-end recognition.
  \item In the D-ST case, our end-to-end trainable TR greatly improves the end-to-end performance of $PRE_s$, $REC_s$ and $F\text{-}score$ by 3.4\%, 5.7\% and 4.6\%, respectively.
  \item The proposed method (D-ST + TR) significantly outperform the {D-BASE}+TR ($\mathcal{L}_{S}$) by 6.6\%.
\end{itemize}
Above results demonstrate the effectiveness of text recommender as well as the whole framework.
\begin{table}[!thbp]
\begin{center}
\scalebox{1}{
\begin{tabular}{|l|cccc|}
\hline
D-BASE      & \checkmark   & \checkmark &                              &               \\
D-ST      &                         &            & \checkmark        & \checkmark    \\ \hline
TR ($\mathcal{L}_{S}$)      & \checkmark   &            & \checkmark        &               \\
TR    &                        & \checkmark &                             & \checkmark    \\
\hline
$PRE_s$     &  69.91  & 72.84 & 64.88  & 68.28 \\
$REC_s$     &  54.34  & 61.73 & 61.54  & 67.21  \\
$F\text{-}score$   &  61.15  & 66.83 & 63.17 & \textbf{67.74}  \\ \hline              
\end{tabular}
} 
\end{center}
\caption{
    The end-to-end evaluation on IC15. 
}
\label{end-our}
\end{table}
\subsection{Comparison with State-of-The-Arts}
\textbf{Comparison on detection.}
We only evaluate the detection results on IC13 because there {are} no results reported on IC15.
From Table. \ref{tab_detection_performance}, we find that the D-BASE already outperforms existing approaches by a large margin thanking to the robust EAST, but still suffers from low recall due to the complicated motion scenarios.
As expected, the spatial-temporal strategy can significantly improves {\emph{REC} by 4\% and \emph{F\text{-}measure} by 1.3\%}, but with a 4.3\% drop on \emph{PRE}.
Actually, boosting recall performance is more important when facing {very} low recall results, which generally results in the precision decreasing.
\begin{table}[h]
\begin{center}
\scalebox{1}{
\begin{tabular}{|l|c|c|c|}
\hline
Methods                                         & $REC$ & $PRE$  & $F\text{-}measure$  \\ \hline
Khare et al. \cite{khare2015new}                 & 41.40 & 47.60 & 44.30 \\
Zhao et al. \cite{zhao2011text}                  & 47.02 & 46.30 & 46.65 \\
Shivakumara \cite{shivakumara2012multioriented} & 53.71 & 51.15 & 50.67 \\
Yin et al. \cite{yin2013robust}                  & 54.73 & 48.62 & 51.56 \\
Wang et al. \cite{wang2018scene}                 & 51.74 & 58.34 & 54.45 \\ \hline
D-BASE                                        & 56.21 & \textbf{85.76} & 67.91 \\
D-ST                                            & \textbf{60.23} & 81.45 & \textbf{69.25} \\
\hline
\end{tabular}
}
\end{center}
\caption{Detection results on IC13.}
\label{tab_detection_performance}
\end{table}

\textbf{Comparison on tracking.}
Table \ref{tab_reid} shows the comparing results on IC13 and IC15.

\emph{Evaluation on IC13}.
T-BASE outperforms the reported results by a large margin 0.35 on $ATA_D$, 0.08 on $MOTP_D$ and 0.37 on $MOTA_D$.
TR can further separately improve the performance by 0.14 and 0.03 on $ATA_D$ and $MOTA_D$, and maintain the $MOTP_D$ performance.

\emph{Evaluation on IC15}.
T-BASE also achieves comparable {results} with previous methods.
Comparing to the best reported result, TR significantly improves the $ATA_D$ and $MOTA_D$ by 0.12 and 0.03, but falls behind \cite{yang2017unified} on $MOTP_D$.
However, \cite{yang2017unified} points out that $ATA_D$ is the most important metric in IC15 because $ATA_D$ measures the tracking performance over all the text.
\begin{table}[h]
\begin{center}
\scalebox{0.9}{
\begin{tabular}{|l|l|c|c|c|}
\hline
Dataset & Methods & $ATA_D$  & $MOTP_D$  & $MOTA_D$ \\
\hline
        &   IC13's base \cite{karatzas2013icdar} &  0.00 & 0.63 & -0.09\\
        &   TextSpotter \cite{neumann2013combining} & 0.12  & 0.67 & 0.27\\
IC13    &   Nguyen et al. \cite{nguyen2014video} &     0.15  & - & -\\ \cline{2-5}
        &   T-BASE&                                0.50 & 0.75 & 0.64 \\
        &   TR &                             \textbf{0.64}  & \textbf{0.75} & \textbf{0.67}\\         
        \hline
        &   Stradvision-1 \cite{karatzas2015icdar} &  0.32 & 0.71 & 0.48\\
        &   Deep2Text-I \cite{karatzas2015icdar} &    0.45 & 0.71 & 0.41\\
        &   Wang et al. \cite{wang2017end} &           0.56 & 0.70 & 0.57\\
IC15    &   Yang et al. \cite{yang2017unified} &       0.61 & \textbf{0.79} & 0.66\\ \cline{2-5}
        &   T-BASE&                                 0.53 & 0.76 & 0.65 \\
        &   TR &                             \textbf{0.65}  & 0.76 & \textbf{0.68}\\                  
\hline
\end{tabular}
}
\end{center}
\caption{Tracking performance evaluation on IC13 and IC15. The suffix `D' means tracking is applied for detection.
}
\label{tab_reid}
\end{table}

\textbf{Comparison on end-to-end evaluation.}
Conventionally, we also place the frame-wise recognition results on IC15 by referring to the previous works, in which we just treat the recognition output of the selected text region as the final recognition result in each text stream.
Table. \ref{end-tradition} shows that this setting also achieves the state of the art.
\begin{table}[!thbp]
\begin{center}
\scalebox{1.0}{
\begin{tabular}{|l|ccc|}
\hline
Method & $MOTP_R$ & $MOTA_R$ & $ATA_R$ \\
\hline
Stradvision \cite{karatzas2015icdar}  & 0.69 & 0.57 & 0.29 \\
Deep2Text \cite{karatzas2015icdar}  & 0.62 & 0.35 & 0.19 \\
Wang et al. \cite{wang2017end}  & 0.70 & 0.69 & 0.60 \\ \hline
Ours  & \textbf{0.76} & \textbf{0.69} & \textbf{0.63} \\ \hline      
\end{tabular}
} 
\end{center}
\caption{
    The traditional end-to-end evaluation on IC15.
    The suffix `R' {means} tracking is applied for measuring recognition.
}
\label{end-tradition}
\end{table}
\begin{figure}[h]
\begin{center}
\includegraphics[width=1.0\linewidth]{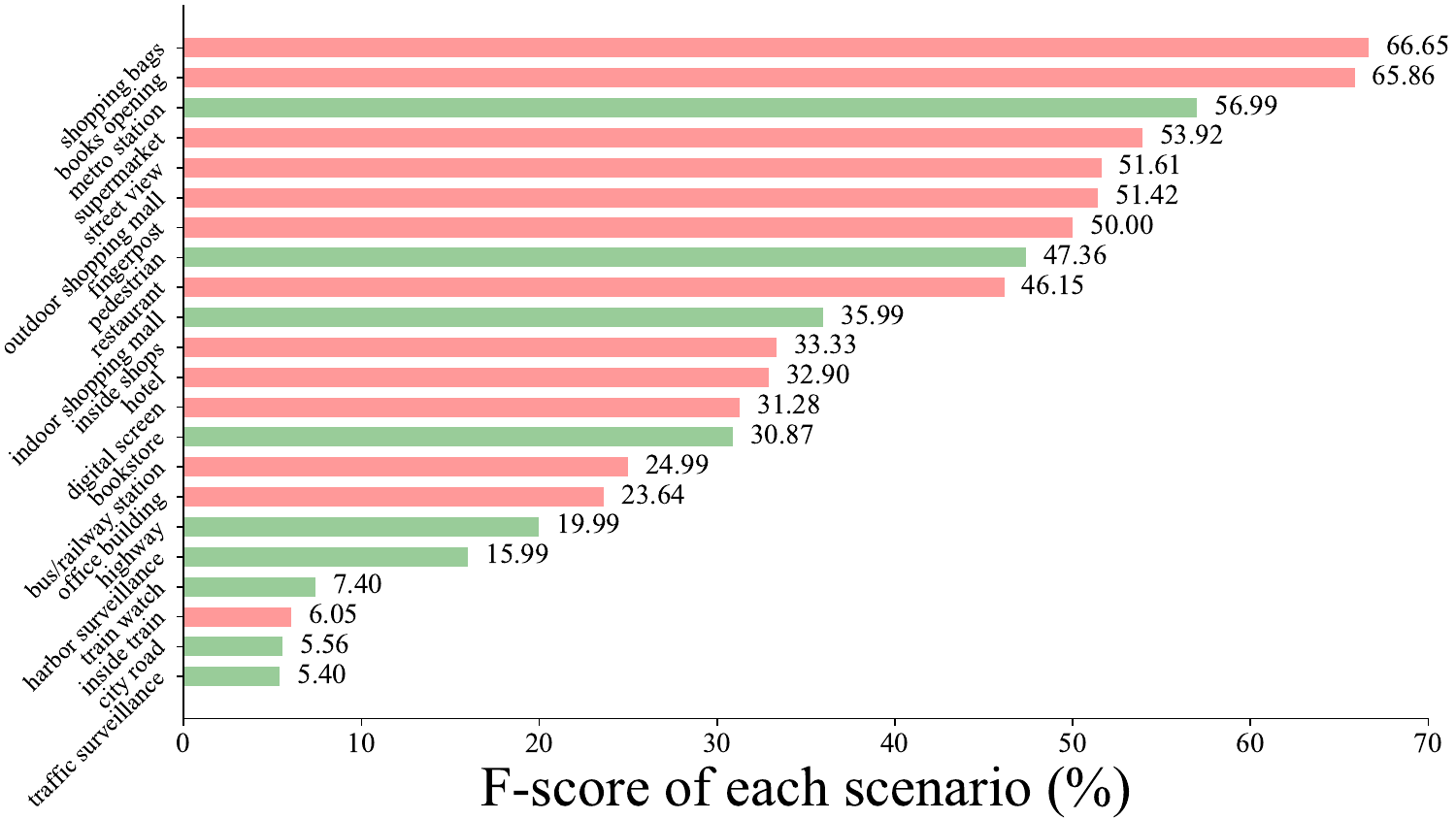}
\vspace{-0.55cm}
\end{center}
   \caption{Illustration of \emph{F-score} performance on 22 scenarios. Indoor/Outdoor scenarios are highlighted in red/green.}
   \vspace{-0.3cm}
\label{fig_challenges}
\end{figure}

\subsection{Challenges on LSVTD}
We also evaluate the proposed framework on the collected LSVTD.
Figure \ref{fig_challenges} shows the \emph{F-score} performance on 22 different scenarios.
In brief, we find that
video text spotting in \emph{books opening} and \emph{metro station} is easier comparing to that in \emph{bookstore} or \emph{shopping mall}.
While video text spotting in \emph{city road}, \emph{camera surveillance} and \emph{train watching} is still very challenging due to extremely complex background and {fierce} shaking, which needs more efforts to boost method's performance.
Overall, spotting text in outdoor is more {challenging} than that in indoor.
More statistics and visualization results {are} shown in {supplementary materials}.

%% file: sections/6_conclu.tex
\section{Conclusion}
In this paper, we propose a fast and robust video text spotting framework named as YORO by integrating a well-designed spatial-temporal video text detector and a novel text recommender.
The video text detector is responsible for recalling more text by referring the relation between different frames.
Moreover, we concentrate on developing the text recommender for selecting the highest-quality text from tracked text streams and then only recognizing the selected text region once, which not only significantly speeds up the recognition process, and also {improves} the video text spotting performance.
In addition, we release a larger-scale video scene text dataset for better evaluating video text spotting algorithms.
In future, we'll further explore the spatial-temporal video text spotting in an end-to-end trainable way.